%% file: PaperForReview.tex
\documentclass[10pt,twocolumn,letterpaper]{article}

 \usepackage{wacv}              

\usepackage{graphicx}
\usepackage{amsmath}
\usepackage{amssymb}
\usepackage{booktabs}
\usepackage{framed,multirow}
\usepackage{multirow}
\usepackage{hhline}
\usepackage{pifont}
\usepackage{booktabs}
\usepackage{amsmath}
\usepackage{amssymb}
\newcommand{\cmark}{\ding{51}}
\newcommand{\xmark}{\ding{55}}
\usepackage{xcolor}
\usepackage{xcolor}
\usepackage{enumitem}
\usepackage{colortbl}
\usepackage{amsfonts}  
\usepackage{algpseudocode} 
\usepackage{graphicx}  
\usepackage{bm}        
\usepackage{caption}   
\usepackage{mathtools} 
\usepackage{multirow}  
\usepackage{xcolor}    
\usepackage{algorithm}
\usepackage{algpseudocode} 
\usepackage{subcaption}
\usepackage{float}
\usepackage[pagebackref,breaklinks,colorlinks]{hyperref}

\usepackage[capitalize]{cleveref}
\crefname{section}{Sec.}{Secs.}
\Crefname{section}{Section}{Sections}
\Crefname{table}{Table}{Tables}
\crefname{table}{Tab.}{Tabs.}
\usepackage{xcolor}

\def\wacvPaperID{819} 
\def\confName{WACV}
\def\confYear{2026}
\definecolor{softlavender}{RGB}{211, 216, 255}  
\definecolor{softblue}{RGB}{136, 176, 245}  

\definecolor{custompurple}{HTML}{9b59b6}
\definecolor{customgreen}{HTML}{2ecc71}
\begin{document}

\title{
{SasMamba: A Lightweight Structure-Aware Stride State Space Model for 3D Human Pose Estimation
}
}

\author{
Hu Cui\textsuperscript{1} \quad 
Wenqiang Hua\textsuperscript{2} \quad  
Renjing Huang\textsuperscript{1} \quad 
Shurui Jia\textsuperscript{1} \quad  
Tessai Hayama\textsuperscript{1}\thanks{Corresponding author.} \quad \\
{\normalsize 
\textsuperscript{1}Information and Management Systems Engineering, Nagaoka University of Technology}\quad \\
{\normalsize 
\textsuperscript{2}School of Computer Science and Technology, Xi’an University of Posts and Telecommunications}\\
{\tt\footnotesize \{s227006, s245028\}@stn.nagaokaut.ac.jp, huawenqiang@xupt.edu.cn
}\\
{\tt\footnotesize 
rjhuang27@gmail.com, 
t-hayama@kjs.nagaokaut.ac.jp
}
}
\maketitle

\begin{abstract}
    
Recently, the Mamba architecture based on State Space Models (SSMs) has gained attention in 3D human pose estimation due to its linear complexity and strong global modeling capability. However, existing SSM-based methods typically apply manually designed scan operations to flatten detected 2D pose sequences into purely temporal sequences, either locally or globally. This approach disrupts the inherent spatial structure of human poses and entangles spatial and temporal features, making it difficult to capture complex pose dependencies.
To address these limitations, we propose the Skeleton Structure-Aware Stride SSM (SAS-SSM), which first employs a structure-aware spatiotemporal convolution to dynamically capture essential local interactions between joints, and then applies a stride-based scan strategy to construct multi-scale global structural representations. This enables flexible modeling of both local and global pose information while maintaining linear computational complexity.
Built upon SAS-SSM, our model SasMamba achieves competitive 3D pose estimation performance with significantly fewer parameters compared to existing hybrid models.
The source code is available at \url{https://hucui2022.github.io/sasmamba_proj/}.
\end{abstract}

  \begin{figure}[t]
    \centering
    \includegraphics[width=0.45\textwidth]{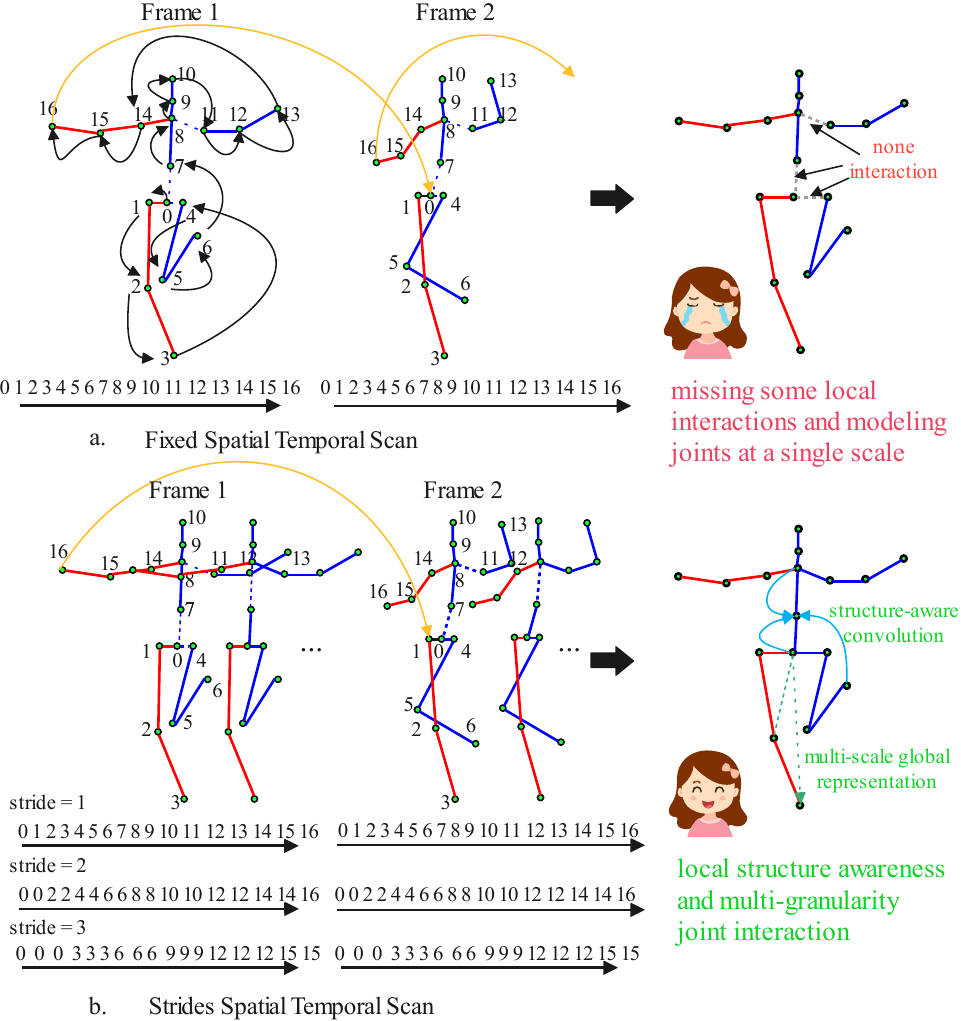}
    \vspace{-2mm}
    \caption{
      Comparison of Pose Sequence Processing: Flattened Scan vs. Structure-Aware Scan.
    }
    \label{fig:Introduction}
    \vspace{-4mm}
  \end{figure}

  \section{Introduction}
Monocular 3D human pose estimation is a fundamental task in computer vision with wide-ranging real-world applications, including augmented and virtual reality \cite{lin2010augmented, orts2016mixed}, autonomous driving \cite{choi2020cars}, and human-computer interaction \cite{shotton2011real, pavlovic1997visual}. Typically, this task involves two separate steps: 2D pose detection to localize keypoints in the image plane, followed by 2D-to-3D lifting to infer joint positions in 3D space. However, due to inherent depth ambiguity and self-occlusion in monocular data, accurately recovering 3D poses from 2D keypoints remains highly challenging.
To address these challenges, deep learning methods have shown significant progress, with Transformer-based architectures, in particular, gaining attention for their ability to model long-range spatial-temporal dependencies.
For example, PoseFormer \cite{zhao2023poseformerv2} introduces a pure Transformer framework that models spatial and temporal dependencies to estimate the 3D pose of the center frame in a video.
MHFormer \cite{li2022mhformer} generates and refines multiple pose hypotheses through a multi-stage Transformer design to better handle depth ambiguity.
MixSTE \cite{zhang2022mixste} alternates spatial and temporal Transformer blocks to separately model inter-joint and per-joint motion correlations for enhanced spatiotemporal representation.
KTPFormer \cite{peng2024ktpformer} enhances self-attention by injecting kinematic and trajectory priors, improving spatial-temporal modeling with minimal overhead.
MotionAGFormer \cite{mehraban2024motionagformer} combines Transformer and GCN branches to jointly capture global and local joint dependencies for better 3D structure learning. Despite their effectiveness, these methods often suffer from high computational and memory costs due to the quadratic complexity of full attention, especially when applied to long sequences of 2D keypoints.

To alleviate this limitation, recent studies \cite{mondal2024hummuss, zhang2025pose, huang2025posemamba, cui2025hgmamba, zheng2025mamba} have explored efficient attention alternatives such as state space models (SSMs)~\cite{gu2022parameterization,guefficiently, gu2021efficiently, gu2023mamba, mehta2022long, liu2024vmamba, zhu2024vision}, which aim to preserve strong global modeling capability while significantly reducing computational overhead.
These architectures provide linear-time sequence modeling and preserve global modeling capacity, enabling scalable and effective pose estimation. For instance,
HumMUSS \cite{mondal2024hummuss} presents a stateful, attention-free SSM-based model enabling efficient real-time 3D pose and mesh prediction with constant memory and time complexity.
PoseMagic \cite{zhang2025pose} proposes a hybrid Mamba-GCN architecture that fuses global spatiotemporal modeling with local joint dependencies for improved accuracy and efficiency.
PoseMamba \cite{huang2025posemamba} introduces a bidirectional global-local SSM block with a spatial reordering strategy to enhance spatiotemporal modeling with linear complexity.
HGMamba \cite{cui2025hgmamba} combines a Shuffle-Mamba stream for global spatiotemporal modeling and a Hyper-GCN stream for local structure modeling through a dual-branch architecture.
While these SSM-based methods demonstrate impressive efficiency and global modeling capabilities, they often rely on flattening spatially structured pose sequences into 1D temporal inputs, which disrupts the human body's spatial topology and entangles temporal and spatial features, hindering the modeling of complex multi-scale dependencies (Figure \ref{fig:Introduction}-a).

To address these limitations, we propose a novel structure-aware SSM framework that preserves spatial topology and captures multi-scale spatiotemporal dependencies in a unified and efficient manner (Figure \ref{fig:Introduction}-b). Specifically, we introduce the Skeleton Structure-Aware Stride SSM (SAS-SSM), which first leverages a structure-aware spatiotemporal convolution to dynamically model local joint interactions while respecting the human body's spatial configuration. Subsequently, a stride-based scan mechanism is employed to construct global representations at multiple scales, enabling long-range dependency modeling without compromising spatial integrity. This dual-stage design allows the model to flexibly capture both fine-grained local structures and coarse global patterns, all within a linear complexity framework.

Built upon the SAS-SSM module, we further develop SasMamba, a compact yet powerful model for 3D human pose estimation. SasMamba achieves competitive performance on standard benchmarks while significantly reducing the parameter count and computational overhead compared to existing hybrid Transformer-GCN or SSM-based architectures. Our framework demonstrates that it is possible to maintain spatial-awareness and multi-scale temporal reasoning simultaneously, without resorting to computationally expensive attention mechanisms or manually designed scan patterns.

\noindent
\textbf{In summary, our contributions are as follows:}
\begin{itemize}
	\item We propose a novel Skeleton Structure-Aware Stride SSM (SAS-SSM) module, which preserves the spatial topology of human poses and captures multi-scale spatiotemporal dependencies through a combination of structure-aware convolution and stride-based scanning.
	
	\item We introduce \textit{SasMamba}, a lightweight and efficient SSM-based model for 3D human pose estimation that achieves strong performance with significantly fewer parameters compared to existing Transformer- or hybrid-based architectures.
	
	\item Our framework unifies local and global modeling in a single pipeline, enabling accurate and scalable 3D pose estimation while maintaining linear computational complexity in both training and inference.
	
	\item Extensive experiments on benchmark datasets (Human3.6M and MPI-INF-3DHP) demonstrate that our method achieves competitive or state-of-the-art results, validating the effectiveness and generalizability of our approach.
\end{itemize}

 \section{Related Work}
 \subsection{3D Human Pose Estimation}

 3D human pose estimation aims to recover the 3D coordinates of human body joints from visual inputs. Existing methods can be broadly categorized by their methodological pipeline. Direct regression approaches predict 3D joint positions directly from raw images using end-to-end CNNs or Transformers~\cite{zhou2019hemlets, chen2021anatomy, li2022mhformer}, requiring large-scale annotations and often facing generalization challenges. In contrast, 2D-3D lifting methods first extract 2D joint keypoints using pre-trained detectors~\cite{newell2016stacked, sun2019hrnet, liu2020single}, then lift them to 3D via neural networks such as fully-connected networks~\cite{martinez2017simple}, graph convolutional networks~\cite{zhao2019semantic, yu2023gla}, or recurrent networks~\cite{hossain2018exploiting}. These decouple visual perception and 3D reasoning, reducing annotation costs but still suffering from depth ambiguity and limited temporal modeling.
 
 More recently, Transformer-based architectures~\cite{li2022mhformer, wang2023temporalformer, zhao2023poseformerv2,zhang2022mixste,peng2024ktpformer,mehraban2024motionagformer} have been adopted for spatiotemporal modeling in 3D pose estimation due to their strong sequence modeling ability. However, Transformers require quadratic computational complexity and massive resources, which hinder their deployment in real-time or long-sequence scenarios. To overcome these limitations, recent studies~\cite{mondal2024hummuss, zheng2025mamba, huang2025posemamba, cui2025hgmamba} have explored attention-free alternatives based on State Space Models (SSMs)~\cite{gu2023mamba, gu2021efficiently, zhu2024vision}. These SSM-based methods achieve linear complexity while preserving strong long-range dependency modeling, making them promising for scalable and efficient 3D HPE.
 However, most of them process pose sequences as flattened temporal inputs, thereby overlooking the inherent skeletal structure of the human body.
 
\subsection{State Space Models}
  \begin{figure*}[h]
    \centering
    \includegraphics[width=0.8\textwidth]{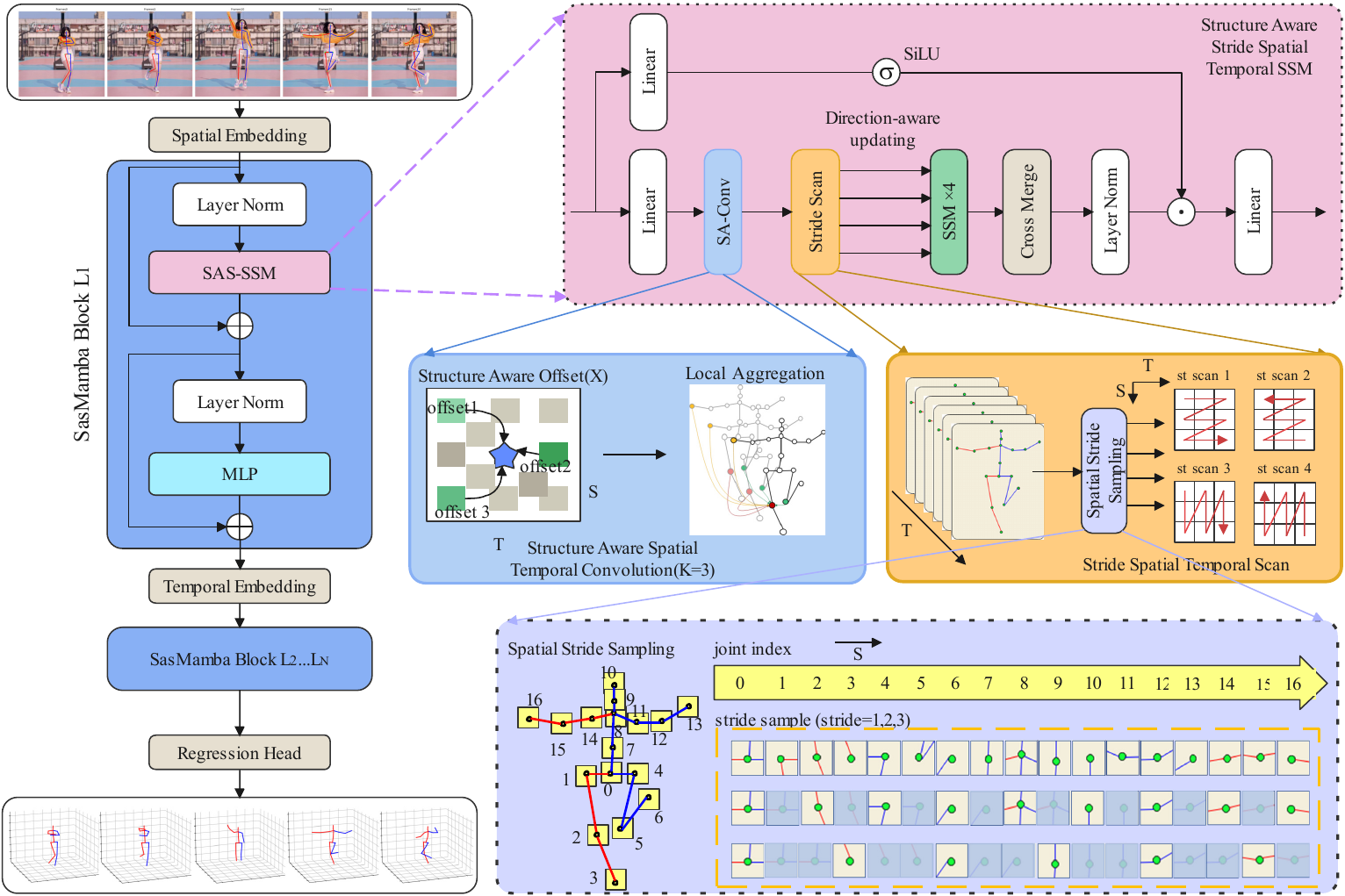}
    \caption{
    \textbf{The overall framework of SasMamba.} The input 2D keypoint sequence is projected into a high-dimensional space, enhanced with positional and temporal embeddings, and processed by SasMamba blocks. To ensure consistent sequence length during Spatial Stride Sampling, invalid tokens (in gray) are replaced with the most recent valid joint token.}
    \label{fig:Overall}
  \end{figure*}
  State Space Models (SSMs) have recently emerged as efficient alternatives to Transformers for sequential modeling, offering linear-time inference and strong long-range dependency modeling. Mamba~\cite{gu2023mamba} introduced a novel parameterization that enables scalable training and has since inspired several vision-oriented variants. Vision Mamba~\cite{liu2024vmamba}, VMamba~\cite{zhu2024vision}, and more recently PlainMamba~\cite{yang2024plainmamba} and DAMamba~\cite{zovo2024damamba} focus on adapting Mamba to visual tasks by addressing a key challenge: preserving spatial structure during sequence modeling. Techniques such as bidirectional or adaptive scanning have been proposed to maintain local patch adjacency and improve spatial continuity, which are crucial for capturing both local and global visual context in image data.
  
  This structural preservation issue is even more critical in 3D human pose estimation, where sparse skeleton data lacks rich semantics and thus depends more heavily on preserving its non-Euclidean topology. However, existing SSM-based methods~\cite{mondal2024hummuss, huang2025posemamba, cui2025hgmamba, zheng2025mamba} largely ignore this issue, flattening 2D joint sequences into 1D temporal inputs and thereby disrupting skeletal structure and mixing spatial-temporal features. To overcome this, our method aims to preserve skeletal topology and enable efficient multi-scale spatiotemporal modeling for accurate and efficient 3D pose estimation.
  

\section{Methodology}
\label{sec:method}
  \subsection{Preliminary}
  State Space Models (SSMs) \cite{gu2021efficiently,gu2023mamba} are a type of sequence modeling technique widely used in deep learning. These models capture dynamic systems by using an intermediate hidden state $h(t) \in \mathbb{R}^N$.
  The system is described by the following equations:
\begin{equation}
h'(t) = A h(t) + B x(t),\quad
y(t) = C h(t),
\end{equation} 
Where the matrices \( A \in \mathbb{R}^{N \times N} \), \( B \in \mathbb{R}^{N \times 1} \), and \( C \in \mathbb{R}^{1 \times N} \) control the system's dynamics and the relationship between the state and the output.

To apply continuous-time models in real-world scenarios, it's need to convert them into a discrete form. One commonly used approach is the Zero-Order Hold (ZOH) method, which assumes that the input remains constant in each time interval. The continuous-time parameters (A, B) are transformed into discrete forms:
\begin{equation}
\begin{aligned}
\overline{A} &= e^{\Delta A}, \\
\overline{B} &= (\Delta A)^{-1}(e^{\Delta A} - I) \Delta B.
\end{aligned} \label{eq:2}
\end{equation}
where \( \Delta \) denotes the time step between samples. The resulting discrete-time model can then be written as:
\begin{equation}
    h_t=\overline{A} h_{t-1}+\overline{B} x_t, \quad y_t=C h_t .
\end{equation}
It enables highly efficient parallel computation by converting the operations into convolutional version:
\begin{equation}
y=x * \overline{K}, \quad \overline{K}=\left(C \overline{B}, C \overline{A B}, \ldots, C \overline{A}^{L-1} \overline{B}\right),
\end{equation}
where \(\overline{K} \in \mathbb{R}^L\) represents the SSM kernel, and \(\ast\) denotes the convolution operation. This parallel structure greatly improves both computational efficiency and scalability.

While traditional SSMs like S4 \cite{gu2021efficiently} achieve linear time complexity, their fixed parameters limit their ability to capture complex sequence dynamics. To address this, the Mamba model \cite{gu2023mamba} introduces a dynamic, input-dependent parameterization. Instead of using fixed transition matrices \(A\) and \(B\), it computes parameters \(B \in \mathbb{R}^{ L \times N}\), \(C \in \mathbb{R}^{ L \times N}\), and \(\Delta \in \mathbb{R}^{ L \times D}\) directly from the input sequence \(x \in \mathbb{R}^{ L \times D}\), enabling more flexible and context-aware modeling.

In Mamba, continuous-time parameters are dynamically adjusted based on the input sequence $x_t$ using selective functions: 
\begin{equation}
    \Delta_t=s_{\Delta}\left(x_t\right), \quad B_t=s_B\left(x_t\right), \quad C_t=s_C\left(x_t\right),
\end{equation}
The input-dependent discrete parameters $\overline{A_t}$ and $\overline{B_t}$ can be computed from Eq. \ref{eq:2}. The discrete state transition and
observation equations are as follows:
\begin{equation}
    h_t=\overline{A_t} h_{t-1}+\overline{B_t} x_t, \quad y_t=C_t h_t .
\end{equation}

\subsection{Skeleton Structure Aware Stride SSM} 

Given the input 2D keypoint feature $X \in \mathbb{R}^{T \times V \times C}$, where $T$ is the number of pose frames, $V$ is the number of human joints, and $C$ denotes the feature dimension (typically $C=2$ for raw 2D coordinates), several existing methods~\cite{mondal2024hummuss, huang2025posemamba, zhang2025pose} reshape the input into flattened sequences such as $X_{st} \in \mathbb{R}^{(TV) \times C}$ or $X_{ts} \in \mathbb{R}^{(VT) \times C}$, mimicking the representation of 2D images. These flattened representations are then processed using State Space Models (SSMs) for bidirectional sequence modeling. 
This flattening process inevitably breaks the skeletal topology and entangles spatial and temporal cues, leading to structural information loss and spatiotemporal confusion. Such disruption undermines the ability to model fine-grained spatial dependencies and temporal dynamics essential for accurate 3D human pose estimation.


In contrast to previous SSM-based approaches, we propose the \textbf{Skeleton Structure-Aware Stride SSM (SAS-SSM)}, as illustrated in Fig.~\ref{fig:Overall}, which explicitly incorporates skeletal structure into the modeling process. Specifically, we first apply a \emph{structure-aware spatiotemporal convolution} to dynamically aggregate local joint features based on the underlying topology, thereby enhancing the model’s ability to extract meaningful semantic representations from the pose structure.
To further enable interactions between joints across varying spatial distances, we introduce a \emph{stride-based sampling strategy} guided by joint indices. This design allows the SSM to capture global contextual information at multiple scales while preserving the original joint ordering, facilitating both structural integrity and efficient long-range dependency modeling.
\paragraph{Structure-aware Spatiotemporal Convolution.}
We first feed the feature map into an $offset(\cdot)$ function to predict the local spatial and temporal offsets $\Delta p \in \mathbb{R}^{ T \times V \times 2}$, for each joint  (for simplicity, one offset for each joint):
\begin{equation}
    \Delta p  = offset(X)
\end{equation}
Then, these offsets, together with the original spatiotemporal  coordinate $p \in \mathbb{R}^{ T \times V \times 2}$, are used to dynamically adjust the spatio-temporal neighboring joints.
\begin{equation}
    p'[t, v, :] = p[t,v,:] + \Delta p[t,v,:]
\end{equation}
Next, we sample the neighboring joints features using the adjusted positions $p'$ employing bilinear interpolation for this process.
\begin{equation}
    X'\left(t, v\right)=\sum_{(n, m)} w_{n m} \cdot X\left(t_n, v_m\right)
\end{equation}
where $\left(t_n, v_m\right)$ are the four neighboring pixels around $\left(t, v\right)$, 
$w_{n m}$ are the bilinear interpolation weights based on the distance between the offset position and its neighboring pixels.
\begin{equation}
    w_{nm} = \left \| p'[t,v,0]-t_n \right \| \cdot  \left \| p'[t,v,1]-v_m \right \|
\end{equation}
After obtaining $K^2$ dynamic neighbor features, we fuse them with the current joint features:
\begin{equation}
    X_{struct}=Conv2d(X) + \sum_{1}^{K^2} {W}_k {X'_{k}} \label{eq-daconv}
\end{equation}
where ${W}_k  \in \mathbb{R}^{C\times C}$ denotes the projection weights of the $k$-th new neighbor joint. $Conv2d(X)$ denotes the local spatio-temporal feature aggregation between the original neighbors. To balance the two, the conv2d convolutional kernel size is set to 3 and K is set to 3. 
\paragraph{Stride-based Sampling and Scan.}
Before using a SSM to extract global features, it is usually necessary to flatten two-dimensional ( $\mathbb{R}^{T \times V}$) data into a one-dimensional sequence ($\mathbb{R}^{TV}$).
However, directly flattening skeletal data may disrupt its inherent topological structure. To mitigate this issue, we propose a stride scan strategy, which performs spatial sampling of skeletal joints at different locations using varying stride values. Since each body part typically contains at least three joints, we set the stride size to \{1, 2, 3\} accordingly (\textcolor{softlavender}{the lower right subplot} of Fig. \ref{fig:Overall}). To ensure parameter stability, we concatenate them along the channel dimension.
\begin{equation}
\begin{split}
x^{1}, x^{2}, x^{3}&= \operatorname{split}(X_{struct} ), \\
x^{1} &\in \mathbb{R}^{T \times V \times \frac{C}{2}}, \quad
x^{2}, x^{3} \in \mathbb{R}^{T \times V \times \frac{C}{4}}
\end{split}
\end{equation}
\begin{equation}
    y^s(t, v,:)= \begin{cases}x^s(t, v,:), & \text { if } v \bmod s=0 \\ x^s(t, v',:), & \text { otherwise, } v^{\prime}=\left\lfloor\frac{v}{s}\right\rfloor \times s\end{cases}
\end{equation}
\begin{equation}
X_{stride} = \operatorname{concat}(y^1, y^2, y^3)
\end{equation}
where $s$ denotes the stride size, $v'$ denotes the skipped joints, which are filled with the values of the preceding valid joints to maintain a consistent sequence length along the joint dimension.

Then, following the strategies in \cite{huang2025posemamba, cui2025hgmamba, zhang2025pose,mondal2024hummuss}, we employ four separate information streams to process the data, enabling the model to capture diverse spatio-temporal patterns and improve its expressive power (\textcolor{orange}{orange subplot} of Fig. 
\ref{fig:Overall}).
\begin{align}
X_{sas} &= \text{SSM}_{\text{temp-fwd}}(X_{\text{stride}}) + \text{SSM}_{\text{temp-bwd}}(X_{\text{stride}}) \notag \\
&\quad + \text{SSM}_{\text{spatial-fwd}}(X_{\text{stride}}) + \text{SSM}_{\text{spatial-bwd}}(X_{\text{stride}})  \label{eq-scandirection}
\end{align}



\subsection{Network Architecture} 
The \textcolor{softblue}{overview} of model architecture is shown in Fig. \ref{fig:Overall}. The model takes 2D input sequence $X^{in}\in \mathbb{R} ^{T \times V \times 2}$ and then maps each joint to a $C$-dimensional feature $X^0 \in \mathbb{R}^{T \times V \times C}$ by a linear projection layer. Then a spatial positional embedding $P^s \in \mathbb{R}^{1\times V \times C}$ and a temporal positional embedding $P^t \in \mathbb{R}^{T \times 1 \times C}$ is added to the tokens before and after the first SasMamba block. 
Subsequently, the skeleton token sequence is processed through $L-1$ SasMamba blocks to capture the underlying 3D structural features. 
The output of block $l$ can be summarized as :
\begin{equation}
\begin{aligned}
    X^{'}_l &=  X_{sas,l}+ X_{l-1} \\
    X_l &= MLP(LN(X^{'}_l)) + X^{'}_l
\end{aligned}
\end{equation}
Finally, a regression head is employed to generate the 3D pose estimates $\hat{P}\in \mathbb{R} ^{T \times V \times 3}$.
The network is trained in an end-to-end manner, the final
loss function $\mathcal{L}$ is defined as:
\begin{equation}
\mathcal{L} = \mathcal{L}_w + \lambda_t \mathcal{L}_t + \lambda_m \mathcal{L}_m,
\end{equation}
where \(\mathcal{L}_w\) is the Weighted Mean Per-Joint Position Error (WMPJPE) \cite{zhang2022mixste}, \(\mathcal{L}_t\) is the Temporal Consistency Loss (TCLoss) \cite{hossain2018exploiting}, and \(\mathcal{L}_m\) is the Mean Per-Joint Velocity Error (MPJVE) \cite{pavllo20193d}. The coefficients \(\lambda_t\) and \(\lambda_m\) are used to control the relative importance of each term.  In our experiments, we set $\lambda_m = 20.0$ and $\lambda_t = 0.5$.


  
\section{Experiments}
We comprehensively evaluate the proposed SasMamba architecture on two large-scale datasets, i.e., Human3.6M \cite{ionescu2013human3} and MPI-INF-3DHP \cite{mehta2017monocular}.
\subsection{Datasets and Evaluation Metrics}
\textbf{Human3.6M} is a standard indoor 3D human pose estimation dataset, featuring 11 subjects performing 15 common actions, resulting in a total of 3.6 million video frames. Following the established protocol \cite{ionescu2013human3, huang2025posemamba, zhang2025pose}, subjects 1, 5, 6, 7, and 8 are used for training, while subjects 9 and 11 serve as the test set. Evaluation employs the Mean Per Joint Position Error (MPJPE) under two protocols: Protocol 1 (P1) measures MPJPE after root joint alignment, and Protocol 2 (P2) uses Procrustes analysis for a rigid alignment before error calculation. 
\\
\textbf{MPI-INF-3DHP} is another large-scale 3D human pose dataset comprising approximately 1.3 million frames captured in diverse environments, including green screen, non-green screen, and outdoor settings. Following previous works \cite{huang2025posemamba,cui2025hgmamba,mehraban2024motionagformer}, we utilize MPJPE as the evaluation metric.

\begin{table*}[t]
	\centering
	\fontsize{9}{10}\selectfont{
		\resizebox{\textwidth}{!}{
			\begin{tabular}{l|c|cc|cccccc}
				\toprule\toprule
				Method   &  Venue   & Seq2Seq& $T$ &Parameter&MACs &MACs/frames &P1(mm) $\downarrow$&P2(mm) $\downarrow$&P1(mm)$^{\dagger}$ $\downarrow$ \\
				\midrule
				
				P-STMO~\cite{shan2022p} &ECCV'22& \xmark&243& 6.2M&0.7G &3M & 42.8&34.4&29.3\\
				STCFormer~\cite{tang20233d}& CVPR'23& \cmark&243& 4.7M&19.6G & 80M& \textbf{41.0}&\textbf{32.0}&22.0\\
				GLA-GCN~\cite{yu2023gla} &ICCV'23& \xmark&243&1.3M &1.5G &6M & 44.4&34.8&\textbf{21.0}\\
				HDFormer~\cite{chen2023hdformer} &IJCAI'23& \cmark&96&3.7M &0.6G &6M & 42.6&\underline{33.1}&{21.6}\\
				MotionAGFormer-XS~\cite{mehraban2024motionagformer} &WACV'24& \cmark&27& 2.2M& 1.0G&37M & 45.1&36.9&28.1\\
				MotionAGFormer-S~\cite{mehraban2024motionagformer} &WACV'24& \cmark&81& 4.8M& 6.6G&81M & 42.5&35.3&26.5\\
				HGMamba-XS~\cite{cui2025hgmamba} &IJCNN'25& \cmark&27& 2.8M& 1.14G&42M & 44.9&38.3&29.5\\
				HGMamba-S~\cite{cui2025hgmamba} &IJCNN'25& \cmark&81& 6.1M& 8.02G&99M & 42.8&35.9&22.9\\
				PoseMamba-S~\cite{huang2025posemamba} &AAAI'25& \cmark&243& 0.9M& 3.6G&15M & 41.8&35.0&22.0$^\star$\\
				\midrule
				\textbf{SasMamba} &-& \cmark&243& 0.64M& 1.3G&5M & \underline{41.48}&34.84&\underline{21.44}\\
				\midrule\midrule   
				MixSTE~\cite{zhang2022mixste} &CVPR'22& \cmark&243& 33.6M& 139.0G&572M & 40.9&{32.6}&21.6\\
				PoseFormerV2~\cite{zhao2023poseformerv2}& CVPR'23& \xmark&243& 14.3M&128.2G & 528M& 45.2&35.6&-\\
				MotionBERT~\cite{zhu2023motionbert} &ICCV'23& \cmark&243&42.3M & 174.8G& 719M& {39.2}&32.9&{17.8}\\
				KTPFormer~\cite{peng2024ktpformer} &CVPR'24& \cmark&243&33.7M & 69.5G& 286M& 40.1&\textbf{31.9}&{19.0}\\
				MotionAGFormer-B~\cite{mehraban2024motionagformer} &WACV'24& \cmark&243&11.7M & 48.3G& 198M& \underline{38.4}&{32.6}&19.4\\
				MotionAGFormer-L~\cite{mehraban2024motionagformer} &WACV'24& \cmark&243&19.0M & 78.3G& 322M& \underline{38.4}&\underline{32.5}&17.4\\
				PoseMamba-B~\cite{huang2025posemamba} &AAAI'25& \cmark&243&3.4M & 13.9G& 57M& 40.8&{34.3}&\underline{16.8} \\
				PoseMamba-L~\cite{huang2025posemamba} &AAAI'25& \cmark&243&6.7M & 27.9G& 115M& \textbf{38.1}&\underline{32.5}&\textbf{15.6} \\
				\midrule
				\textbf{SasMamba-large} &-& \cmark&243&4.1M & 8.56G& 35M& 39.77&33.61&20.92\\
				\bottomrule\bottomrule
			\end{tabular}
		}
	}
	\caption{Quantitative comparisons on Human3.6M dataset. $T$ is the number of input frames. Seq2seq refers to estimating 3D pose sequences rather than only the center frame. MACs/frames represents multiply-accumulate operations for each output frame. P1: MPJPE error
		(mm). P2: P-MPJPE error (mm). P1$^\dagger$: P1 error on 2D ground truth. The best result is shown in bold, and the second-best result is underlined. $^\star$ denotes results reproduced by our implementation using the official code. The upper part lists \textbf{small/lightweight models} focusing on efficiency and fast training, while the lower part shows \textbf{larger models} mainly for reference to overall performance.
	}
	\label{tab:human36mainresults}
\end{table*}

\subsection{Implementation Details}

Our base model, \textbf{SasMamba}, consists of $L = 10$ blocks with a hidden dimension of $D = 64$. A larger variant, \textbf{SasMamba-L}, is constructed by simply doubling both the number of blocks and the hidden dimension.
Our implementation is based on PyTorch 2.0.0 with Python 3.8, running on Ubuntu 20.04. We use CUDA 11.8 for GPU acceleration. All training is conducted on a single NVIDIA RTX 4090 GPU, while inference and evaluation are performed on a single NVIDIA A6000 GPU.
During training, we utilize both \textit{ground-truth 2D poses} and \textit{pre-estimated 2D poses} generated by the Stacked Hourglass model~\cite{newell2016stacked}. The model is trained for 150 epochs with a mini-batch size of 16, using the AdamW optimizer with a weight decay of 0.01. The initial learning rate is set to $5 \times 10^{-4}$ and decays exponentially by a factor of 0.99 at each epoch.
For the \textbf{Human3.6M} dataset, we follow~\cite{mehraban2024motionagformer, zhu2023motionbert, huang2025posemamba} and use both ground-truth and Stacked Hourglass 2D poses. For the \textbf{MPI-INF-3DHP} dataset, only ground-truth 2D poses are used, in accordance with standard evaluation protocols~\cite{tang20233d, zhao2023poseformerv2}.

\subsection{Performance comparison on Human3.6M}

We present a comprehensive comparison of our proposed SasMamba and SasMamba-large models against recent state-of-the-art methods on the Human3.6M dataset. To ensure fairness, we only consider models that do not rely on additional pre-training data beyond the dataset itself. As summarized in Table~\ref{tab:human36mainresults}, our SasMamba model achieves a P1 error of 41.48mm and a ground truth-based P1$^\dagger$ error of 21.44mm, while requiring only 0.64M parameters and 1.3G MACs in total. Notably, this represents one of the most lightweight models among recent approaches, yet still delivers competitive accuracy, outperforming several recent compact models such as PoseMamba-S \cite{huang2025posemamba}, MotionAGFormer-S \cite{mehraban2024motionagformer} and HGMamba-S \cite{cui2025hgmamba}.

Furthermore, our SasMamba-large variant achieves comparable accuracy to larger transformer-based models (e.g., MotionBERT \cite{zhu2023motionbert}, KTPFormer \cite{peng2024ktpformer} and MotionAGFormer-B~\cite{mehraban2024motionagformer}) with far fewer parameters and computations. This is mainly due to its structure-aware design and the Mamba-based state space architecture, which capture skeletal topology and long-range temporal dependencies more efficiently than attention.
In addition, per-action comparisons in Table~\ref{tab:human36mex2} reveal that our model consistently performs better than other lightweight counterparts on several challenging actions, such as \textit{Greet}, \textit{Purchase}, and \textit{Smoke}. This further demonstrates the effectiveness and generalizability of our approach in capturing subtle motion variations.


%

\begin{table*}[ht]\Large
	\centering
	\resizebox{\linewidth}{!}{
		\begin{tabular}{lc|ccccccccccccccc|c}
			\toprule\toprule
			\textbf{MPJPE(P1)} & $T$ & Dire. & Disc. & Eat & Greet & Phone & Photo & Pose & Purch. & Sit & SitD & Smoke & Wait & WalkD & Walk & WalkT & Avg\\
			\midrule
			MixSTE~\cite{zhang2022mixste}CVPR'22 	& 243 &\textbf{37.6} &\underline{40.9} &{\textbf{37.3}}& 39.7 &{42.3} &{49.9} &40.1 &39.8 & 51.7 &\textbf{55.0} & 42.1 & 39.8 & 41.0 &{27.9} &\underline{27.9} & 40.9\\
			P-STMO~\cite{shan2022p}ECCV'22 							& 243 &38.9 & 42.7 				& 40.4 & 41.1 			 & 45.6 & {49.7} 		& 40.9 & 39.9 & 55.5 & 59.4 & 44.9 & 42.2 & 42.7 & 29.4 & 29.4 & 42.8\\
			Einfalt~\textit{et al.}~\cite{einfalt2023uplift}WACV'23 & 351 & 39.6 			& 43.8 				& 40.2 & 42.4 			 & 46.5 & 53.9 					& 42.3 & 42.5 & 55.7 & 62.3 & 45.1 & 43.0 & 44.7 & 30.1 & 30.8 & 44.2\\
			STCFormer~\cite{tang20233d}CVPR'23 						& 243 & 39.6 			& 41.6 				&\underline{37.4} & 38.8 & 43.1 & 51.1 					& {39.1} & 39.7 & \underline{51.4} & 57.4 &\underline{41.8} &{38.5} & 40.7 &\textbf{27.1} & 28.6 & 41.0\\
			HDFormer~\cite{chen2023hdformer}IJCAI'23						& 96 & \underline{38.1} &43.1 &39.3& 39.4 &44.3& \textbf{49.1}& 41.3& 40.8&53.1 &62.1 &43.3 &41.8 &43.1& 31.0& 29.7& 42.6\\
			PoseMamba~\cite{huang2025posemamba}AAAI'2025    &243 &{38.8} &\textbf{40.8} &38.8 &\underline{35.2} &\underline{42.1} &50.8 &{\textbf{38.8}} &\underline{36.4} &51.8 &61.9 &{42.0} &\textbf{38.4} &\underline{38.7} &28.1 &28.7 &\underline{40.8} \\
			\midrule
			\textbf{SasMamba}								& 243 & 39.02 &43.03 &39.24 &36.12 &43.17& 51.26& 40.00 &37.21 &52.70 &61.74 &43.11 &38.88  &39.84&28.32& 28.60 & 41.48\\
			\textbf{SasMamba-large}		& 243 & 38.41 &41.02 &38.01& \textbf{34.58}& \textbf{41.33}& \underline{49.32}& \underline{38.84} &\textbf{36.12} &\textbf{50.73} &\underline{56.17}& \textbf{41.59} &\underline{38.44} &\textbf{37.22}&\underline{27.19} & \textbf{27.62}& \textbf{39.77}\\
			\midrule\midrule
			
			
			\textbf{P-MPJPE(P2)} 					  				  	& $T$ & Dire.& Disc. & Eat & Greet & Phone & Photo & Pose & Purch. & Sit & SitD & Smoke & Wait & WalkD & Walk & WalkT & Avg\\
			\midrule
			P-STMO~\cite{shan2022p}ECCV'22			  				& 243 &{31.3} & 35.2 & 32.9 & 33.9 & 35.4 &{39.3} & 32.5 & 31.5 & 44.6 &{48.2} &36.3 & 32.9 & 34.4 &{23.8} &{23.9} &34.4\\
			MixSTE~\cite{zhang2022mixste}CVPR'22 	& 243  &{30.8} &\textbf{33.1}&\textbf{30.3}& 31.8& \underline{33.1}& 39.1& 31.1 &\underline{30.5} &42.5& \textbf{44.5} &\underline{34.0} &\textbf{30.8}& 32.7 &\underline{22.1} &\underline{22.9} & \underline{32.6}\\
			Einfalt~\textit{et al.}~\cite{einfalt2023uplift}WACV'23 & 351 & 32.7 & 36.1 & 33.4 & 36.0 & 36.1 & 42.0 & 33.3 & 33.1 & 45.4 & 50.7 & 37.0 & 34.1 & 35.9 & 24.4 & 25.4 & 35.7\\
			STCFormer~\cite{tang20233d}CVPR'23 						& 243 & \textbf{29.5} &\underline{33.2} &\underline{30.6} &31.0 &\textbf{33.0}& \underline{38.0}& \textbf{30.4} &\textbf{29.4}& \underline{41.8}& \underline{45.2}& \textbf{33.6}& 29.5& \textbf{31.6}& \textbf{21.3}& \textbf{22.6}& \textbf{32.0}\\
			HDFormer~\cite{chen2023hdformer}IJCAI'23						& 96 & \underline{29.6}& 33.8& 31.7 &31.3 &33.7& \textbf{37.7}& \underline{30.6}& 31.0& \textbf{41.4}& 47.6 &35.0 &\underline{30.9} &33.7& 25.3 &23.6& 33.1\\
			PoseMamba~\cite{huang2025posemamba}AAAI'2025  			& 243 &32.3 &34.0 &33.2 &\underline{30.2} &{34.9} &40.6 &{32.0} &{31.0} &44.5 &53.0 &36.3 &{31.3} &{33.5} &{23.8} &24.6 &{34.3} \\
			\midrule
			\textbf{SasMamba}									    	  	& 243 & 32.28& 35.70 &33.52& 30.76 &36.35& 40.23& 32.78& 32.07& 43.96 &52.70 &37.04 &31.68  &34.30 &24.35&24.83 &34.84\\
			\textbf{SasMamba-large}							& 243 & {31.87}& 33.95 &32.43& \textbf{29.79} &34.65 &39.15 &31.66 &31.25 &43.15& 49.54 &36.23& 31.08& \underline{32.15}& 23.45 &23.87& 33.61\\
			\bottomrule\bottomrule
		\end{tabular}
	}
	\caption{Quantitative comparisons for each action on Human3.6M using detected 2-D pose sequence obtained by Stacked Hourglass \cite{newell2016stacked}. \(T\) represents the input time frame. The best results are highlighted in bold, and the second-best results are underlined.}
	\label{tab:human36mex2}
\end{table*}

\subsection{Performance comparison on MPI-INF-3DHP}
\begin{table}[h]
  \centering
\label{tab:3dhp}
\fontsize{10}{11}\selectfont{
\resizebox{0.5\textwidth}{!}{
    \begin{tabular}{l|cccc}
    \toprule\toprule
 Method&Venue&$T$ &Seq2Seq&P1(mm) $\downarrow$\\
\midrule
    MixSTE~\cite{zhang2022mixste}& CVPR'22     & 27 &\cmark &54.9\\
    P-STMO~\cite{shan2022p}& ECCV'22     & 81 &\xmark &32.2\\
    STCFormer~\cite{tang20233d}& CVPR'23     & 81 &\cmark&23.1\\
    PoseFormerV2~\cite{zhao2023poseformerv2} &CVPR'23     & 81&\cmark&27.8\\
    GLA-GCN~\cite{yu2023gla} &ICCV'23     & 81 &\xmark &27.8\\
    MotionAGFormer-XS~\cite{mehraban2024motionagformer}& WACV'24     & 27 &\cmark &{19.2}\\
    MotionAGFormer~\cite{mehraban2024motionagformer}& WACV'24     & 81 &\cmark &\underline{18.2}\\
    
    \midrule
    \textbf{SasMamba} & -& 27&\cmark&20.16\\
    \textbf{SasMamba} & -& 81&\cmark&\textbf{17.97}\\
\bottomrule\bottomrule
    \end{tabular}
    }
    }
\caption{Quantitative comparisons on MPI-INF-3DHP. T: Number of input frames. The best and second-best scores are in bold and underlined, respectively.}
\label{tab:3dhpex}
\end{table}

We evaluate our model on the MPI-INF-3DHP dataset and report results for both 27-frame and 81-frame input settings. As shown in Table~\ref{tab:3dhpex}, our 81-frame variant of SasMamba achieves a P1 error of {17.97mm}, outperforming all previous methods including MotionAGFormer-B (18.2mm) and MotionAGFormer-XS (19.2mm) \cite{mehraban2024motionagformer}. Even with a shorter 27-frame input, our SasMamba achieves a competitive result of 20.16mm, surpassing most prior approaches with longer input sequences. 

These results demonstrate the effectiveness and robustness of our model on in-the-wild 3D pose estimation, especially under limited temporal input or real-world variability.

\subsection{Ablation Studies}
\begin{table}
\centering
\resizebox{0.48\textwidth}{!}{
\begin{tabular}{l|cc|cc|l} \toprule\toprule
Component & \multicolumn{2}{c|}{Structure Aware Module} & \multicolumn{2}{c|}{Scan Approach}           & \multirow{2}{*}{P1(mm)}  \\ \cline{1-5}
Methods   & GCN~ & SA-Conv                               & Local-Global Scan & Stride Scan &                         \\ \midrule
1         &    &                                       & \cmark            &             &     43.41                    \\
2         & \cmark     &                                       & \cmark                  &             &  43.26                      \\
3         &      &   \cmark                                    & \cmark                  &             &  42.89                       \\
4         &      &                                       &                   & \cmark            &  42.08                       \\
5         & \cmark     &                                       &                   & \cmark            &  42.16                       \\
6         &      &  \cmark                                     &                   &   \cmark          &  \textbf{41.48}                       \\ \bottomrule\bottomrule
\end{tabular}}
\caption{Impact of Structure-Aware Convolution and Stride Scan in SAS-SSM.}
\label{tab:component}
\end{table}



\begin{table}
\centering
\resizebox{0.4\textwidth}{!}{
\begin{tabular}{lllc|l} \toprule \toprule
K & Params. & Flops. &Epoch Time (sec)& P1(mm)  \\ \hline
1 & 0.417M & 1.74G  &492&   42.67      \\
3 & 0.624M & 2.60G  &672&   \textbf{41.48}      \\ 
5 & 2.01M  & 8.35G  &1115&  43.49       \\
7 & 6.56M  & 27.13G &1582&  43.18       \\ \bottomrule \bottomrule
\end{tabular}}
\caption{Effect of kernel size \(K\) in SA-Conv on model efficiency and accuracy.}
\label{tab:saconv_kernal}
\end{table}


\begin{table}[h]
	\centering
	\resizebox{0.48\textwidth}{!}{
		\fontsize{10}{11}\selectfont{
			\begin{tabular}{lcc|c}
				\toprule \toprule
				ScanDirection & T & Params & P1(mm) $\downarrow$ \\
				\midrule
				S-f: Spatial-forward   &243&0.624M&43.39\\
				S-b: Spatial-backward  &243&0.624M&43.24\\
				S-fb: Spatial-bidirectional &243&0.624M&42.37\\
				T-f: Temporal-forward  &243&0.624M&43.27\\
				T-b: Temporal-backward &243&0.624M&43.23\\
				T-fb: Temporal-bidirectional &243&0.624M&42.11\\
				ST-fb: Spatial+Temporal bidirectional &243&0.624M&\textbf{41.48}\\
				\bottomrule \bottomrule 
	\end{tabular}}}
	\caption{Ablation study on different scan directions in the Stride Scan module.}
	\label{tab:proxy}
\end{table}

For a more in-depth analysis of our SasMamba, we further conduct a series of ablation studies on Human3.6M dataset using the Stacked Hourglass estimated 2D poses as input.
\paragraph{Ablation Study on SAS-SSM Components.}
In Table~\ref{tab:component}, we compare different variants of SAS-SSM to assess the contributions of the Structure-Aware and Stride Scan modules. Compared to conventional GCN-based modeling (Row 3 vs. Row 2), our structure-aware spatiotemporal convolution (SA-Conv) achieves improved performance (42.89mm vs. 43.26mm) without relying on manually defined adjacency matrices, thereby enhancing model flexibility and scalability.
Notably, to preserve the overall lightweight design of our framework, we do not compare against more expressive yet substantially heavier GCN variants such as CTR-GCN or TD-GCN~\cite{chen2021channel, cui2025dstsa, liu2023temporal}.

In terms of the scan strategy, the Stride Scan (Rows 4–6) consistently outperforms the manually defined Local-Global Scan strategy (Rows 1–3) \cite{huang2025posemamba}. Our Stride Scan avoids dataset-specific local index design and naturally adapts to diverse skeletal structures, making it simpler and more effective (e.g., 41.48mm vs. 42.89mm).

These results demonstrate that each proposed component contributes positively to the overall performance while keeping the model lightweight and efficient.
\paragraph{Effect of Kernel Size \(K\) in SA-Conv.}
We investigate the impact of different kernel sizes \(K\) in the structure-aware convolution defined in Equation~\ref{eq-daconv}, where \(K^2\) determines how many joints each joint is allowed to interact with. Table~\ref{tab:saconv_kernal} shows the results of varying \(K\) in terms of model parameters, FLOPs, training time per epoch, and the 3D pose estimation error (P1 in mm).

When \(K=1\), the model achieves reasonable accuracy (42.67mm) with the smallest parameter count (0.417M) and fastest training speed. Increasing the kernel size to \(K=3\) yields the best performance (41.48mm), with only a moderate increase in parameters (0.624M) and training time. However, further increasing \(K\) to 5 or 7 leads to significant growth in both model size and computational cost (e.g., 6.56M params and 27.13G FLOPs for \(K=7\)) while degrading performance. This is likely due to the over-smoothing effect or redundant interactions, which may dilute meaningful local structural information.

These results demonstrate that our method benefits from moderate joint interactions (e.g., \(K=3\)) and avoids unnecessary overhead or performance degradation caused by excessive connectivity.

\paragraph{Effect of Scan Direction.}
We evaluate the effectiveness of different scan directions in the proposed Stride Scan module, as formulated in Equation~\ref{eq-scandirection}. Table~\ref{tab:proxy} compares individual temporal or spatial directions (e.g., T-f, S-b) as well as their combinations.

Using both forward and backward scanning (e.g., S-fb, T-fb) improves performance over unidirectional counterparts, showing that bidirectional modeling provides richer temporal or spatial context. Notably, combining both temporal and spatial bidirectional scans (ST-fb) achieves the best result (41.48mm), demonstrating the advantage of jointly modeling global structure in both dimensions.

\subsection{Qualitative Analysis}
\begin{figure}[t]
    \centering
    \includegraphics[width=0.48\textwidth]{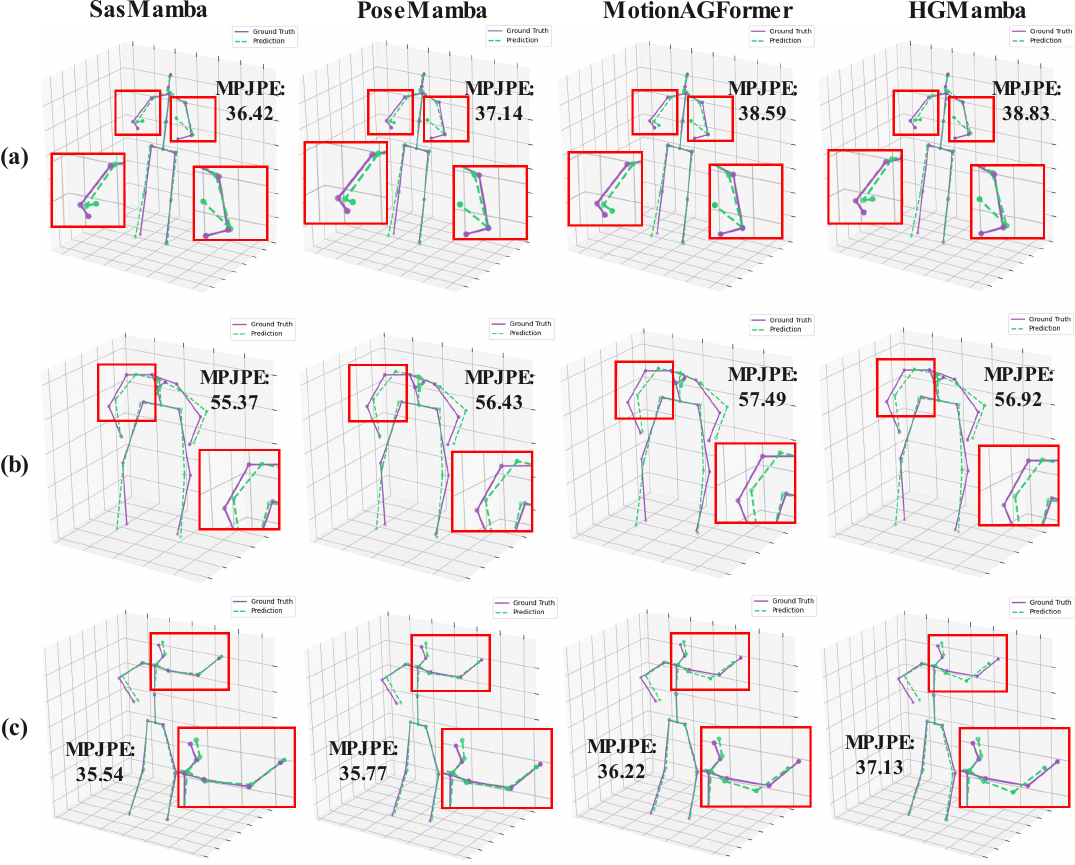}
    \caption{
    	Qualitative comparisons of our proposed {SasMamba} with PoseMamba-S \cite{huang2025posemamba}, MotionAGFormer-S \cite{mehraban2024motionagformer}, and HGMamba-S \cite{cui2025hgmamba} on 3D human pose estimation. The \textcolor{custompurple}{solid purple skeletons} represent the ground-truth 3D poses, while the \textcolor{customgreen}{dashed green skeletons} indicate the predicted 3D poses.
    }
    \label{fig:vis1}
  \end{figure}

\begin{figure}[h]
    \centering
    \includegraphics[width=0.48\textwidth]{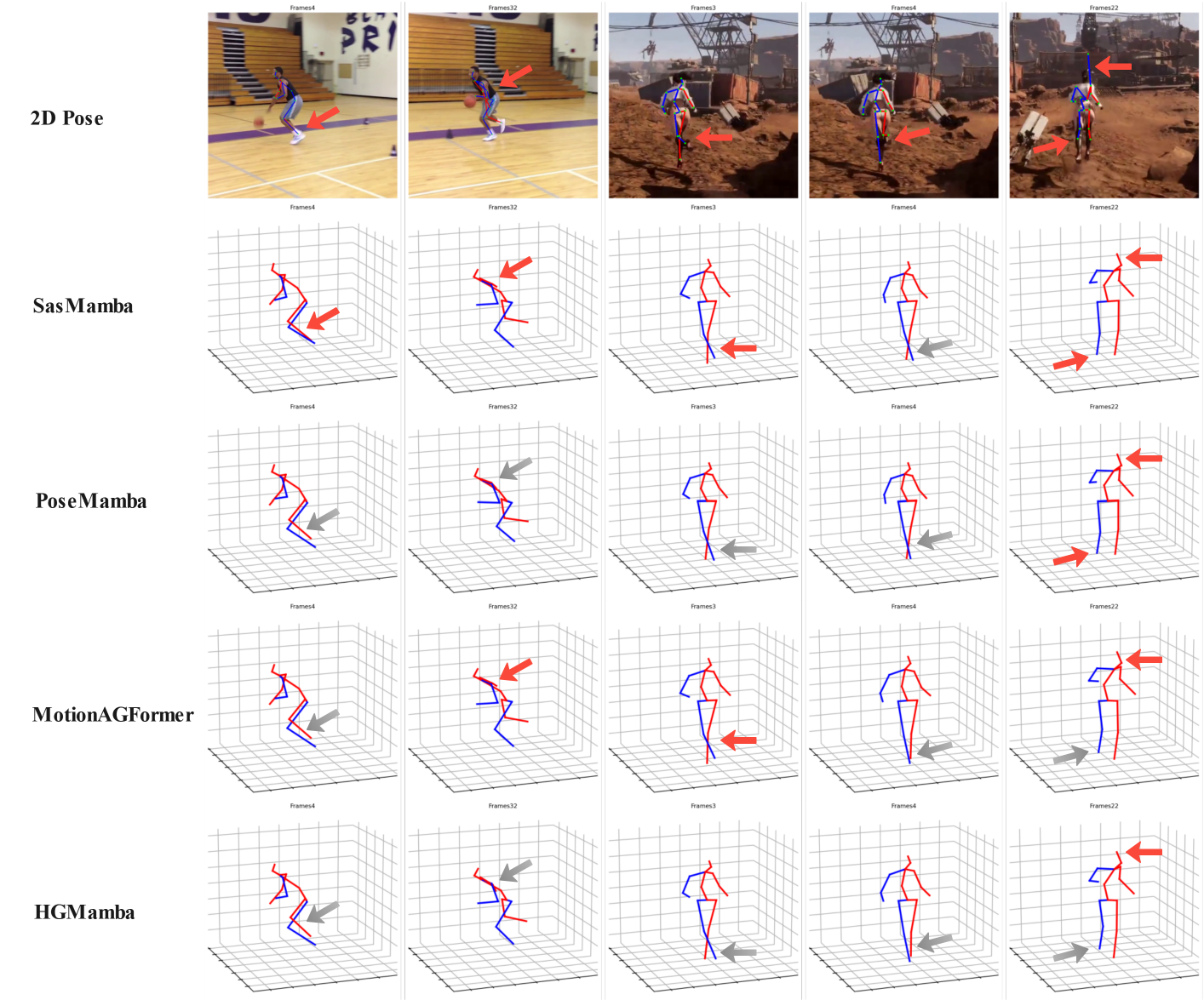}
    \caption{
    	Qualitative comparisons with PoseMamba-S \cite{huang2025posemamba}, MotionAGFormer-S \cite{mehraban2024motionagformer}, and HGMamba-S \cite{cui2025hgmamba} on challenging in-the-wild videos. 
    	\textcolor{red}{Red arrows} indicate accurate estimations, while \textcolor{gray}{gray arrows} highlight unsatisfactory estimations.
    }
    
    \label{fig:vis2}
  \end{figure}
\noindent In Figure~\ref{fig:vis1}, the examples are randomly selected from the evaluation set of Human3.6M. 
When compared with recent lightweight methods, it is clearly observed that SasMamba not only achieves lower overall estimation errors but also reconstructs joint-level structural details more accurately.

In addition, Figure~\ref{fig:vis2} presents in-the-wild pose videos involving both real and synthetic humans, where the 2D poses are obtained using HRNet. 
It can be observed that our method accurately estimates 3D poses for all samples except the one in the fourth column. 
More extensive visualizations and detailed frame-by-frame comparisons can be found in the \textcolor{blue}{supplementary material}
and on our 
\textcolor{blue}{\href{https://imaginative-semolina-67b9c9.netlify.app/}{project page}}.

\section{Conclusion}
We proposed SasMamba, a lightweight framework for monocular 3D human pose estimation built upon the Skeleton Structure-Aware Stride SSM (SAS-SSM) module. By preserving skeletal topology and capturing multi-scale spatiotemporal dependencies through structure-aware convolution and stride-based scanning, SasMamba addresses the limitations of prior SSM methods that flatten pose sequences.
Our model achieves competitive performance on Human3.6M and MPI-INF-3DHP with significantly lower computational cost and fewer parameters than Transformer- or hybrid-based baselines, demonstrating its effectiveness and scalability.  
{\small
\bibliographystyle{ieeenat_fullname}
\bibliography{egbib}
}

\clearpage  
\section*{Supplementary Material}
\input{suppl.tex}

\end{document}

%% file: suppl.tex
This supplementary material provides extended qualitative analyses to complement the main paper. 
Our additional experiments are organized into two categories:  

\begin{itemize}
	\item \textbf{Robustness to Noisy 2D Inputs:}  
	We evaluate SasMamba on mildly challenging wild videos, where erroneous 2D pose detections are present.  
	We further test the model under fixed background and virtual character/scene conditions to analyze its stability.  
	
	\item \textbf{Performance on Challenging Poses:}  
	We investigate SasMamba’s ability to handle extreme human poses.  
	Specifically, we provide qualitative results on aerial rotations and analyze its robustness under different viewpoints, including top and side views.  
\end{itemize}
\section{Robustness to Noisy 2D Inputs}
\begin{figure*}[h]
	\centering
	\includegraphics[]{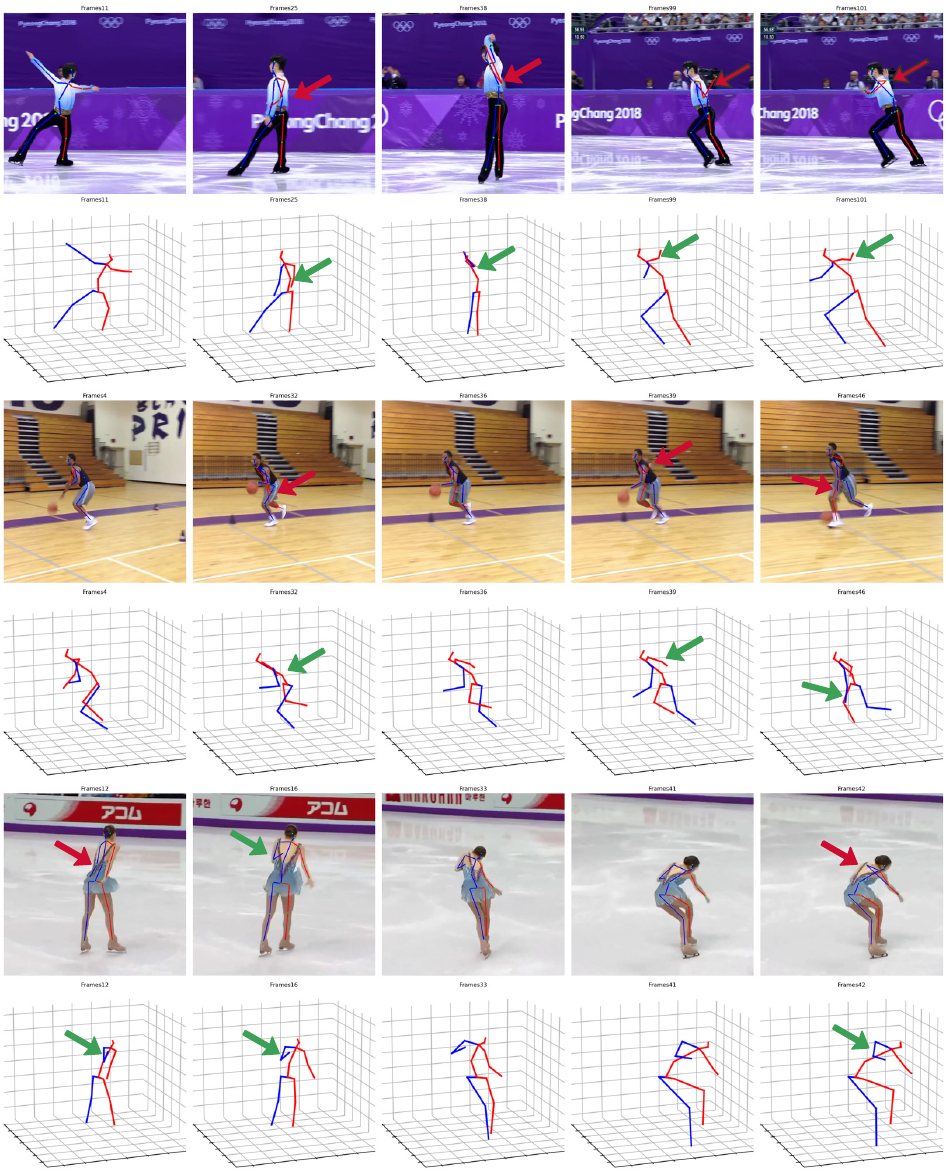}
	\vspace{-2mm}
	\caption{
		Qualitative Results on Mildly Challenging Wild Videos. \textcolor{red}{Red arrows} highlight erroneous 2D pose estimations, 
		while \textcolor{green}{green arrows} indicate correct 3D predictions.
	}
	\label{fig:sup1}
	\vspace{-4mm}
\end{figure*}

To evaluate the robustness and generalizability of SasMamba, we conducted experiments on several moderately challenging, previously unseen real-world videos (Figure~\ref{fig:sup1}). In these videos, human actions evolve smoothly and scene changes occur gradually. Our results show that even when the input 2D poses exhibit noticeable noise (\textcolor{red}{red arrows}), SasMamba consistently produces accurate 3D pose estimates (\textcolor{green}{green arrows}). This demonstrates the strong resilience of the model in handling abrupt or noisy motions within 2D input sequences. 

\begin{figure*}[h]
	\centering
	\includegraphics[width=\textwidth]{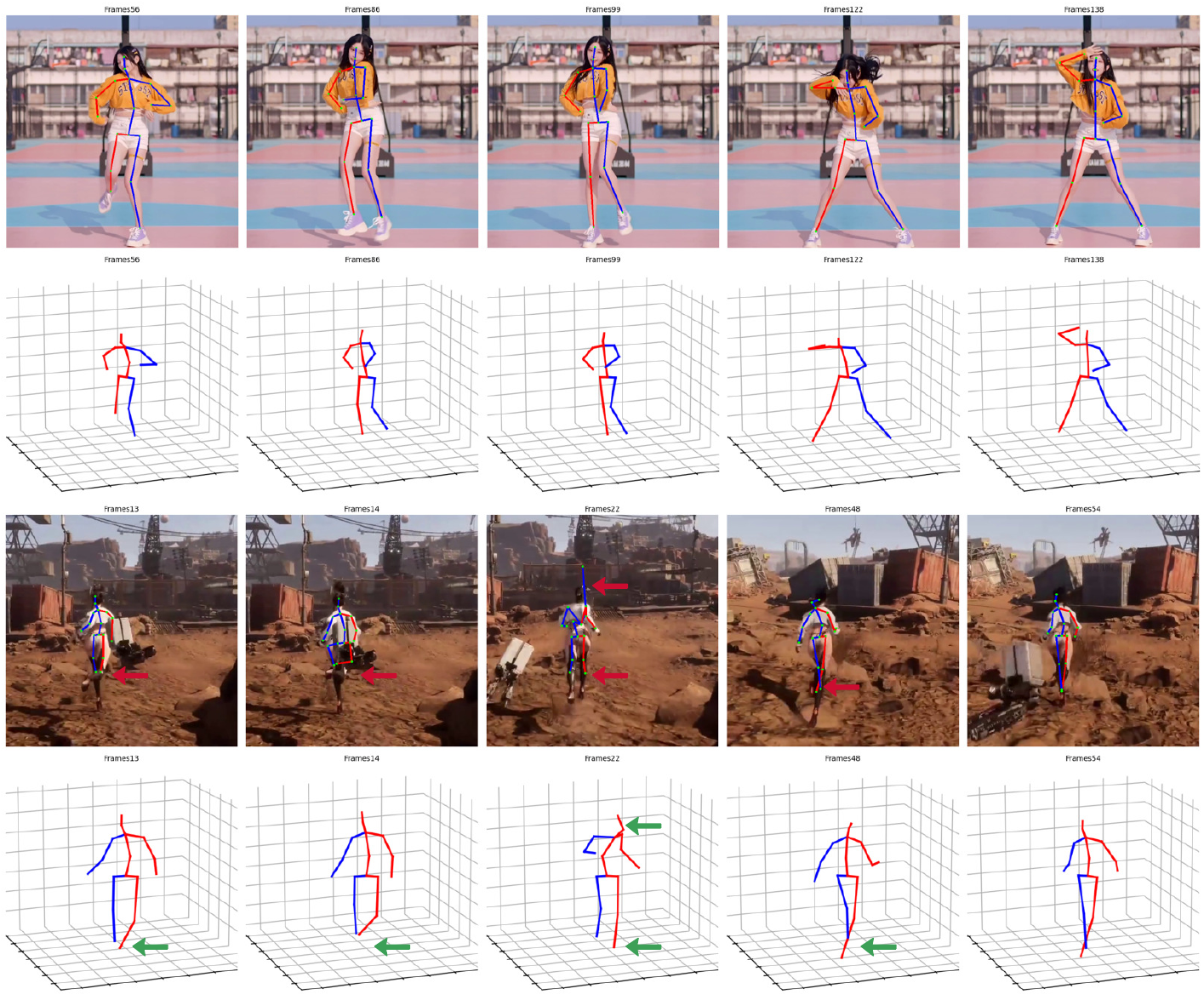}
	\vspace{-2mm}
	\caption{
		Qualitative Results of SasMamba under Fixed Background and Virtual Character/Scene Conditions.
	}
	\label{fig:sup2}
	\vspace{-4mm}
\end{figure*}

Furthermore, Figure~\ref{fig:sup2} illustrates that when backgrounds are fixed or blurred, improvements in the quality of input 2D sequences lead to more stable 3D predictions. In the case of virtual characters performing high-speed motions in synthetic environments, SasMamba remains robust: despite degraded 2D pose quality, the model leverages its global structure-awareness to correct local pose estimation errors effectively.

\section{Performance on Challenging Poses}
\begin{figure*}[h]
	\centering
	\includegraphics[width=\textwidth]{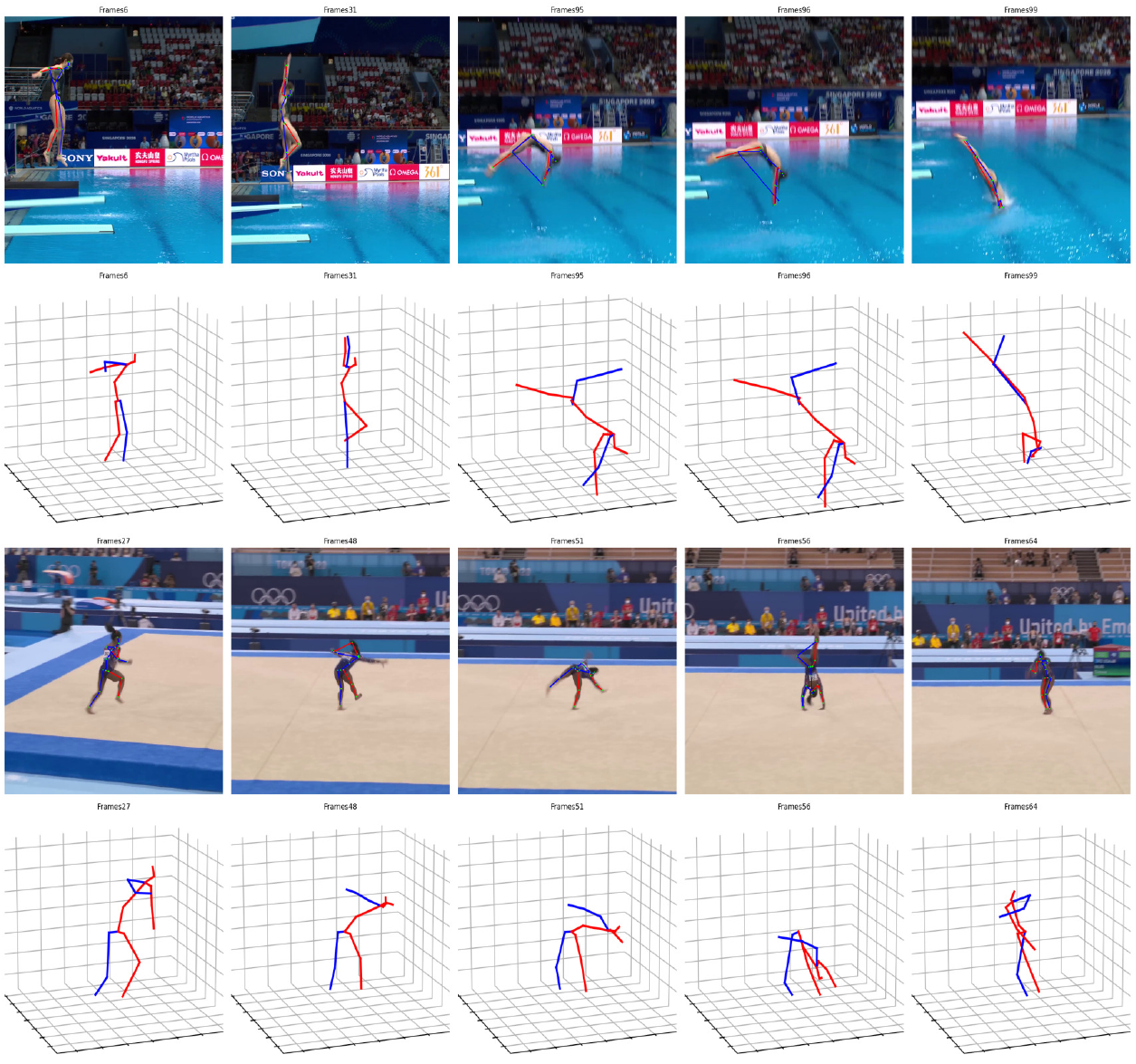}
	\vspace{-2mm}
	\caption{
		Qualitative Results of SasMamba on Extreme Poses (Aerial Rotation).
	}
	\label{fig:sup3}
	\vspace{-4mm}
\end{figure*}

To further explore the robustness limits of SasMamba, we evaluated the model on more extreme motion samples, including diving, gymnastics, and skiing. Unlike the smoother motions in earlier experiments, these actions involve rapid velocity changes and frequent high-difficulty aerial rotations. As shown in Figure~\ref{fig:sup3}, such irregular short-term motion patterns not only pose challenges to 2D detectors—resulting in severely degraded input quality—but also exceed the correction capacity of our model. These findings highlight that, in the 3D pose estimation pipeline, the reliability of the 2D detection stage is as critical as the lifting stage.

\begin{figure*}[h]
	\centering
	\includegraphics[width=\textwidth]{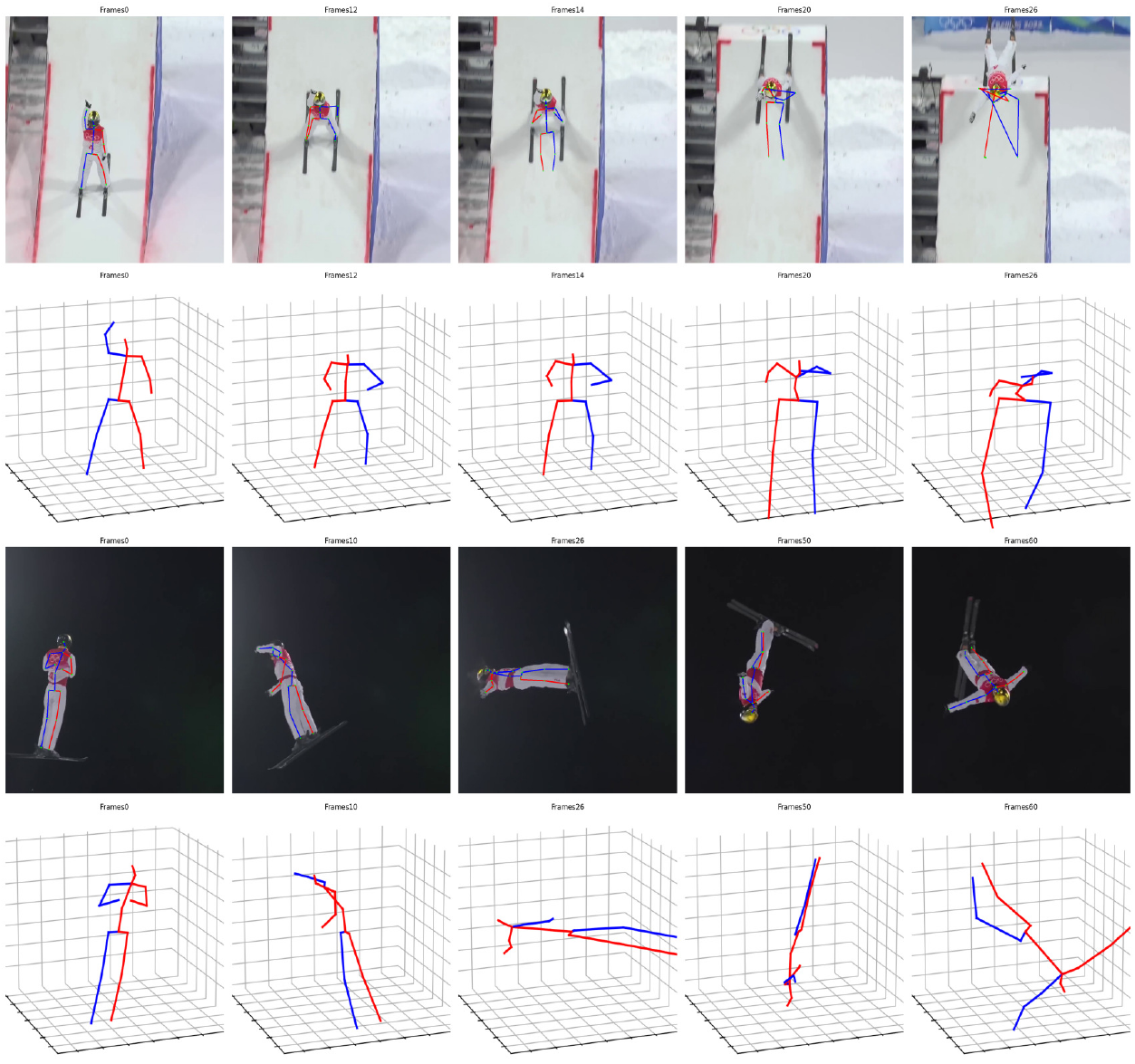}
	\vspace{-2mm}
	\caption{
		Qualitative Results of SasMamba on Aerial Rotations from Top and Side Views.
	}
	\label{fig:sup4}
	\vspace{-4mm}
\end{figure*}

Additionally, in Figure~\ref{fig:sup4}, we evaluated extreme motions of the same subject from two side views. The results indicate that a single, consistent background and lateral viewpoints yield more reliable predictions, whereas top-down views introduce additional challenges due to occlusions and entanglement among joint positions. This further reinforces the observation that, in 3D pose estimation, the performance of the 2D pose detection stage is equally critical as that of the lifting stage.

